\PassOptionsToPackage{sort&compress}{natbib}

\documentclass{article}

\usepackage[margin=1.1in]{geometry}

\usepackage[utf8]{inputenc} 
\usepackage[T1]{fontenc}    
\usepackage{url}            
\usepackage{booktabs}       
\usepackage{amsfonts}       
\usepackage{nicefrac}       
\usepackage{microtype}      
\usepackage{bbm}
\usepackage{amsmath}

\usepackage{array, multirow}

\usepackage{subcaption}
\usepackage{tabularx,ragged2e,booktabs,caption}

\usepackage{graphicx}

\newtheorem{theorem}{Theorem}
\newtheorem{defn}{Definition}

\title{Parity Partition Coding for Sharp Multi-Label Classification}

\author{
  Christopher G. Blake*  \ \ Giuseppe Castiglione* \ \  Christopher Srinivasa \ \  Marcus Brubaker
  \\ *Shared first authorship}

\begin{document}

\maketitle

\begin{abstract}
The problem of efficiently training and evaluating image classifiers that can distinguish between a large number of object categories is considered. A novel metric, sharpness, is proposed which is defined as the fraction of object categories that are above a threshold accuracy. To estimate sharpness (along with a confidence value), a technique called fraction-accurate estimation is introduced which samples categories and samples instances from these categories. In addition, a technique called parity partition coding, a special type of error correcting output code, is introduced, increasing sharpness, while reducing the multi-class problem to a multi-label one with exponentially fewer outputs. We demonstrate that this approach outperforms the baseline model for both MultiMNIST and CelebA, while requiring fewer parameters and exceeding state of the art accuracy on individual labels.
\end{abstract}

\section{Introduction}
Mulit-label classification is the problem of, given an image or instance, accurately labelling its attributes. For example, given an image of a face, a multi-label classification task may require us to identify if the face is male, if there are blue eyes, if the image is smiling, and so on. Since it's possible that each of the attributes can vary independently, the number of possible classes to which an instance can belong can grow exponentially in the number of attributes. A naive and unrealistic approach to this problem is to create a classifier for each of the $2^K$ possible categories (where $K$ is the number of attribute labels). In practice this is not done; and the more sensible approach is to train a binary classifier for each binary attribute, and estimating the instance's category from the outputs of these classifiers (this is the multi-class to binary technique introduced in \cite{AllweinReducingMultiClassToBinary}).

There are three engineering challenges with this approach which we address in this paper: the first is, we want to make sure that the classifier is accurate. Secondly, we want the accurate classifiers to be compact and have a small training time. Finally, we want to be able to test and benchmark such classifiers, and we want to do it in a way that ensures that the classifiers are not biased. In this paper we introduce a technique called \emph{parity partition coding}, a special type of error correcting output code \cite{DietterichErrorCorrectinOutputCodes}. We demonstrate that using this technique results in more accurate classifiers when ensembling a comparable number of models. The technique involves learning both primitive attributes (those with labels), and derived attributes (in our case, attributes that can be viewed as parity functions of primitive attributes). As a sub-technique, for training the derived attributes, we introduce \emph{quadratic feature transformation} which we show decreases training time for learning parity attributes. Finally, we introduce a technique called \emph{fraction accurate estimation} to estimate the fraction of categories that are accurate above a threshold by sampling over categories and then sampling instances of each of these categories. 

We demonstrate our techniques on two problems: a synthetic dataset we call multi-MNIST, and a facial attribute recognition challenge based on the celebA dataset. For the celebA problem, we show that our technique compares to state of the art accuracy for the accuracy of individual attributes. When accuracy is defined as the number of instances in which \emph{all} the attributes must be accurate for the classifier to be accurate, our technique outperforms our baseline ensemble of primitive attribute models.


\section{Prior Work}

The authors of \cite{abbeWainwrightTutorial} argue that there is a relationship between error control coding and machine learning, which inspires our work. Our work herein is in the category of ensemble methods in machine learning, and \cite{DietterichEnsembleMethods} provides a good overview. Our paper also involves learning attributes as in \cite{lampertBetwenClassAttributeAnimalAttributes, liuRecognizingHumanActionsByAttributes}.

The technique we consider today can be viewed as a special case of an error correcting output code invented by Dietterich \emph{et al.} \cite{DietterichErrorCorrectinOutputCodes}. This is a technique used for multi-class classification problems, like digit recognition. The Dietterich technique involves selecting binary attributes that characterize target categories. Each category is then associated with a unique binary attribute string encoding the attributes associated with this category. A sample instance is then fed into each binary categorizer, producing an estimate of the binary attributes of that instance. The category is then chosen to be the category with attribute string closest in Hamming distance to the estimated attribute string. In our work, we consider the problem where attributes are all labelled, and the object categories under considerations are those that are defined by the unique configuration of these attributes. This requires us to train both primitive attribute classifiers \emph{as well as} derived attribute classifiers.

Shannon \cite{Shannon} introduced the concept of error control coding by proving that there exist codes with asymptotic rates above $0$ with error probability that approaches $0$ for channels with independent noise. In the channel coding problem, the fundamental cost is in channel uses. In the attribute classification problem, the analogous costs are training energy, and the energy used for classification of a new instance using the already trained network. In the channel coding case, high rate codes imply high channel use efficiency. For parity partition coding, high rates imply energy efficient training and deployment.

The first paper to recognize the equivalence of the output of a classifier being the output of a noisy channel is \cite{machineLearningMeets}. In \cite{liMultiLabelEmotionClassification}, error control output codes are considered for the multi-label classification problem. The authors of \cite{Hansen1990NeuralNetworksAccurateAndDiverse} consider the problem of ensembling neural networks and points out that these ensembles work if the models are accurate (better than random) and diverse (different models make errors on different inputs). This notion is a weaker notion than our pure independence assumption which informs the theory in Section~\ref{sec:complexityAnalysis}. 

An analysis of the effectiveness of error correcting output codes is in \cite{Masulli2000EffectivenessOE} and in \cite{GuruswamimultiClassLearning}, the authors consider a similar model to ours. We build upon this literature by performing experiments to show what we consider the main advantage of error correcting output codes: an asymptotic savings in the number of models needed in an ensemble to reach high accuracy.



 \section{Problem Setup}
\vspace{-2ex}

\begin{figure} 
\centering
\begin{subfigure}{.15\textwidth}
  \centering
  \includegraphics[width=\textwidth]{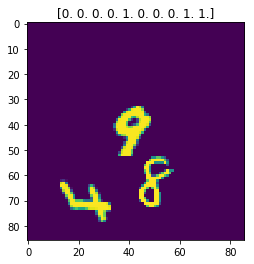}
  \caption{}
  \label{fig:sfig1}
\end{subfigure}
\begin{subfigure}{.15\textwidth}
  \centering
  \includegraphics[width=\textwidth]{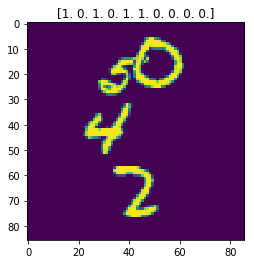}
  \caption{}
  \label{fig:sfig2}
\end{subfigure}
\begin{subfigure}{.15\textwidth}
  \centering
  \includegraphics[width=\textwidth]{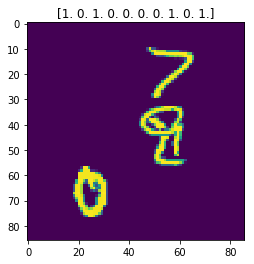}
  \caption{}
 
\end{subfigure}
\caption{Samples from the MultiMnist Dataset, with corresponding attribute strings labelled at the top of each image. This dataset is produced by randomly generating positions and numbers, and then drawing a random MNIST image for each number and placing it at a random position.}\label{fig:multiMNISTExample}
\end{figure}

We will define our problem generally and then use a synthetic multiMNIST dataset to give a concrete example. Example instances of multiMNIST are given in Figure~\ref{fig:multiMNISTExample}. We also consider the attribute labelled celebA dataset, which includes images of celebrity faces with attributes like `has bangs' and `glasses.' We let the  \emph{instance space} be the set of all possible inputs into the classifier \cite{valiantTheoryOfLearnable}.
\vspace{-1ex}
\begin{defn} A binary attribute is a \emph{bipartition} of the instance space. A \emph{primitive attribute} is an attribute that is labelled. A \emph{derived attribute} is an attribute induced by a function of the primitive attributes of an instance. 
\end{defn}
\vspace{-1ex}
For example, in multiMNIST, '1' and '2' are primitive attributes. In celebA, "has bangs" is a primitive attribute. The attribute '1 XOR 2' is a derived attribute for the multiMNIST dataset. This is the set of instances with a '1' but not '2' and a '2' but not '1.' 
\vspace{-1ex}
\begin{defn} An \emph{attribute template} is an ordered tuple of attributes. 
The \emph{primitive attribute string}  of an instance is a string of symbols representing the state of attributes for that instance.
\end{defn}
\vspace{-1ex}
The idea here is that an image encodes a ``message'' within the state of its attributes. The goal of classification is to figure out the primitive attribute string of the instance. For example, in the multi-MNIST problem, the goal is to produce a length $10$ binary string where each element of the string corresponds to the presence ($1$) or absence ($0$) of the $10$ MNIST digits that may be present in the image. See the labels at the top of each image in Figure~\ref{fig:multiMNISTExample} to see examples of primitive attribute strings.

 We let the length of each element in the attribute space (or equivalently the length of the attribute template) be denoted by the symbol $K$. We denote a primitive attribute string of length $K$ as $X_1,\ldots X_K$.


Consider a sequence of $N$ functions, each mapping a primitive attribute string  to $\{0, 1\}$. Let's denote these functions $f_1 f_2 \ldots f_N$. 
\vspace{-1ex}
\begin{defn}
A \emph{length $N$ binary linear error correcting output code} is a sequence of $N$ such functions where each function is a $\mod 2$ sum of a subset of the elements of the primitive attribute string. Note that the sum of these functions may simply output a primitive attribute (\emph{i.e.}, they are the sum of single attribute). We shall also consider derived attributes where $f_i(X_1,\ldots X_K)$ is a $\mod 2$ sum of a subset of the $K$ elements of the primitive attribute string. The functions $f_1 f_2 \ldots f_N$ are called the \emph{encoding functions}.
\end{defn}
\vspace{-1ex}
Without loss of generality, we denote the outputs of the $N$ functions as: $ Y_1, \ldots, Y_N $.  Note that each of these functions induces a partition of the instance space, one partition for each function $f_i$ in the natural way. Precisely, a function $f_i$ mapping an attribute string to $0$ or $1$ bipartitions the instance space into $2$ sets, $A_0$ and $A_1$, where $A_i$ is the set of instances whose attribute string maps to the symbol $i$. \emph{Thus, for each of these partitions we can train a binary classifier, producing $N$ binary classifiers}. This is the key idea of parity partition coding.

After feeding the sample instance into the $N$ classifiers, a \emph{sample output string} is produced, which is not necessarily the \emph{ground truth output string}, which is the string produced when feeding the primitive attribute string into the encoding functions. The task is to find the attribute string with a ground truth output string  which is ``closest'' to the sample output string. Closeness may be measured in Hamming distance.  Estimating the closest ground truth output string is called \emph{decoding}.

The technique of training primitive and parity attribute classifiers and then decoding the produced outputs on an instance we call \emph{parity partition coding}. Note that this is a special case of error correcting output codes \cite{DietterichErrorCorrectinOutputCodes} except in this case we distinguish primitive and parity attributes. 

\vspace{-2ex}
\section{Complexity Analysis and Comparison to Error Control Coding}\label{sec:complexityAnalysis}
\vspace{-1ex}

\begin{figure} 
	\centering
	\includegraphics[width=3.8 in]{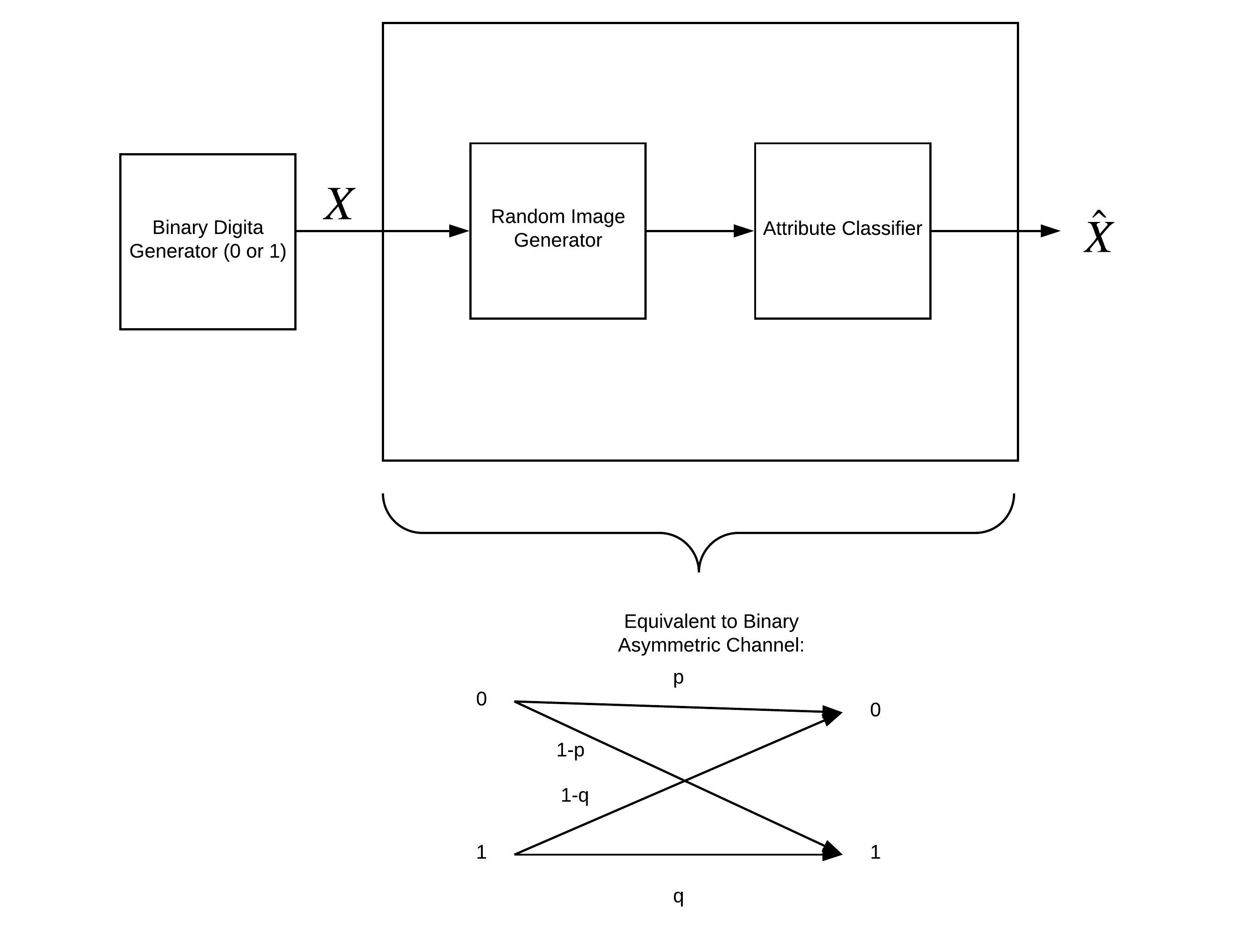}
	\caption{A diagram of a binary classifier and it's equivalence to a binary asymmetric channel. The random image generator takes in as input a binary variable indicating the desired state of a given attribute, and the image generator produces an image that has that attribute. The purpose of the classifier is to classify images created by this random image generator. Sometimes the random image generator will be some randomized function, but it can also be the natural images that arise in an application. We see at the bottom of the graph that once the random distribution and the classifier is fixed, then the labelled black box is exactly equivalent to a binary asymmetric channel, where $1-p$ is the probability an image with attribute label $0$ is classified with a $1$, and $1-q$ is the probability an image with attribute label $1$ is classified with a $0$. When there are multiple attributes (shown in Figure~\ref{fig:attributePartition}, this can be viewed as multiple parallel binary asymmetric channels.}
\end{figure}

\begin{figure} 
	\centering
	\includegraphics[width=6.2 in]{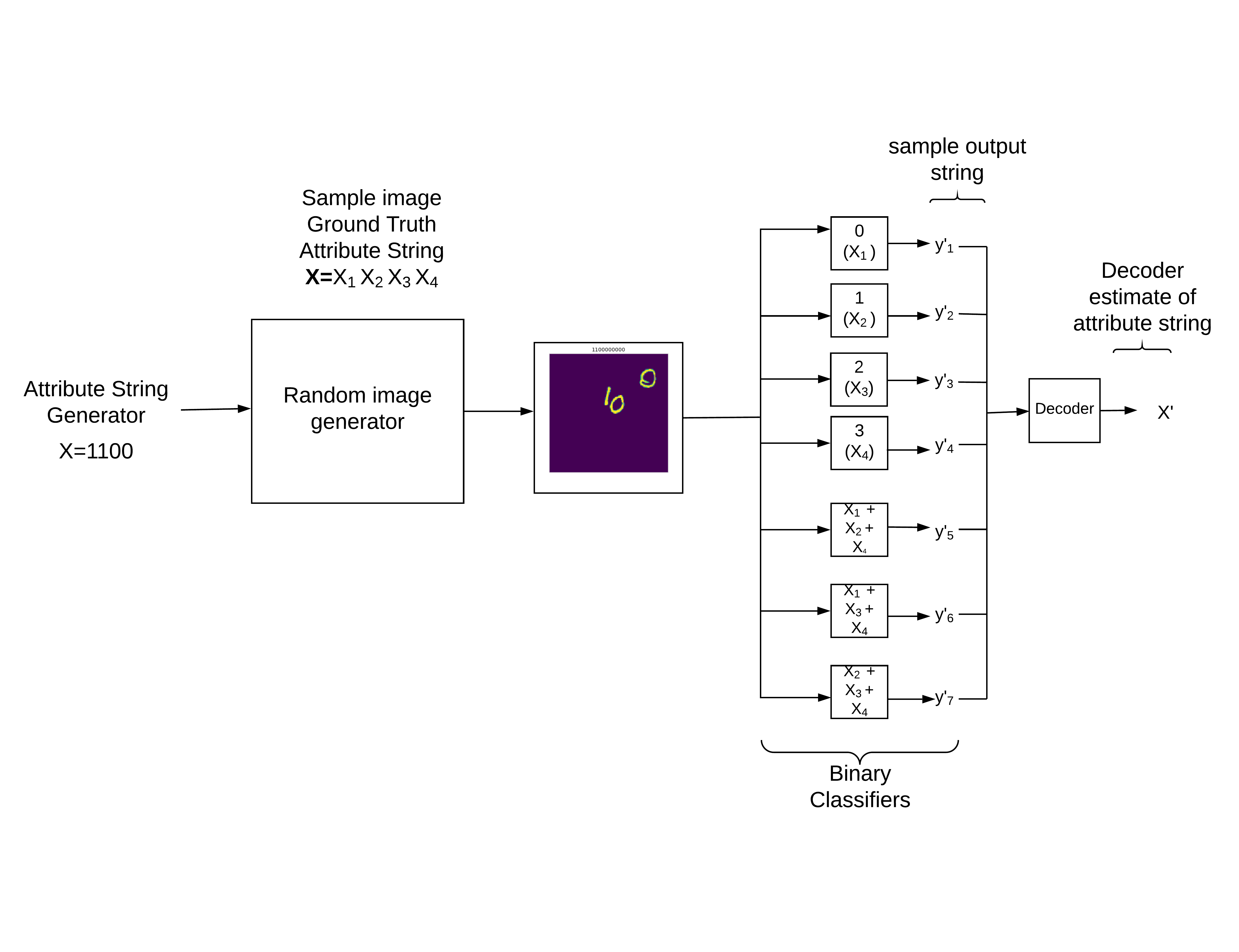}
	\caption{ Diagram for an attribute classifier for classifying 0, 1, 2, and 3 for the multiMNIST dataset using an attribute partition coding scheme based on a $(7, 4)$-Hamming code \cite{hamming}. The random image generator takes in as input a binary string and produces an image with attributes encoded by this binary string. The purpose of the binary classifier and decoding layer is to estimate what the original attribute string was. In the binary classifier layer, the top $4$ classifiers are binary classifiers for $\{0, 1, 2, 3\}$, and the three bottom classifiers are parity partition classifiers: that is, they are classifiers trained to output $1$ when the $\mod 2$ sum of the check bits is $1$. The ground truth symbol for each of the top $4$ attributes is symbolized as $x_1, x_2, x_3, x_4$ as labelled in the diagram.  So, for example, for the bottom parity classifier, this will be a classifier trained to recognize if, for a given image, $x_2+x_3+x_4 = 1$, where addition is performed modulo $2$.} \label{fig:attributePartition}
\end{figure}

We shall use the admittedly overly simple assumption that each binary classifier is a binary asymmetric channel that is independently distributed, where the input is a label for an attribute, and the output is a $1$ or $0$ indicating the classifier's estimate of the state of that attribute for a random instance of that attribute. An illustration of this concept is given in Figure~\ref{fig:attributePartition}. Moreover, we assume we will be testing our classifier on an \emph{independent attribute distribution}. That is, we shall assume that each attribute occurs independently and randomly with probability $1/2$.

We consider error control coding for the binary asymmetric channel. We assume $K$ bits encoded to length $N_{s}=K/R$, where $R$ is the rate of the code. We define error probability as the probability that our classifier does not produce the correct primitive attribute string for an instance.
We consider using a Shannon code \cite{Shannon} that has error probability for the channel $e^{-c_1N_s}$ from some $c_1>0$. (Such codes exist, see \cite{forneyConcatenated, HassaniAlishahiUrbanke}. Moreover they have close to optimal encoding and decoding complexity \cite{blakePhDThesis}).

As a comparison, we consider a repetition code that has length $N_r=K n_r$ where $n_r$ is the number of repetitions. Such a code is equivalent to training $n_r$ separate $K$-output primitive attribute classifiers. A simple probability analysis shows that such a code has error probability that scales as $e^{-c_2 n_r}$ for some $c_2>0$. Thus, to get equivalent error probability for the repetition code as for the Shannon code, it must be that:
\vspace{-0.8ex}
\[
e^{-c_2 n_r}=e^{-c_1 N}
\]
and thus $n_r=c'N$ for another $c'>0$. This means that the length of the repetition code must be $N_r=c'K N$, that is, a $K$ factor greater than the equivalent Shannon code. This is the primary advantage of the parity partition coding technique: an asymptotic savings in the number of binary classifiers needed. We find that training binary classifiers for the derived attributes (those corresponding to parity functions) is more difficult than primitive attributes. This has been observed in the literature \cite{Nye18} when learning a simple parity function (and not a parity of attributes). This motivates the following techniques which we test on multiMNIST: \emph{quadratic feature transformation}, \emph{targeted bagging}, and \emph{pre-trained weight initialization}. We describe each of these technique below and explain how our theory informs how these techniques can help in training.
\vspace{-3ex}

\section{Parity Partition Coding Training Techniques}
\vspace{-1.5ex}
Below we describe the three special techniques we studied for training the parity partition coders. Their use is informed by our theory, which we explain in each subsection. 
\vspace{-1.5ex}
\subsection{Quadratic Feature Transformation}
\vspace{-1.5ex}
 To employ the parity partition coding technique, we need to learn a parity function of primitive attributes. We also want models that learn different parity attributes to be accurate and diverse. Thus, inspired by the technique of quadratic transformation of inputs to a parity function learner \cite{Piazza92artificialneural}, in the output layer of the neural network model, we apply a quadratic transformation of the features before being fed into the final linear layer, via an outer product. Due to symmetry, we retain only the off-diagonal, upper-triangular portion of the outer-product matrix. This transformation makes XOR linearly separable, and thus easier to learn. In our experiments, we found that all other things being equal, quadratic feature transformation cut down the number of training epochs needed by over $50\%$ to reach the same accuracy ($3$ epochs compared to $7$).
\subsection{Targeted Bagging}
\vspace{-1.5ex}
   Adapted from \cite{Breiman1996BaggingP}, \emph{targeted bagging} trains different targets on different splits of the dataset. This leads to decorrelation of the outputs. Roughly speaking, this corresponds to making the outputs of each of the models more independent, which according to our theory should allow for an increase in the effectiveness of the parity partition coding technique, since it makes the errors more independent. In Table~\ref{tab:sharpness_mnist} we can see the effectiveness of this technique.
\vspace{-1.5ex}
\subsection{Pretrained weight initialization}
\vspace{-1.5ex}
We also study a technique called \emph{pre-trained weight initialization}. This involves re-using feature extractors trained for a separate classification task, and holds them fixed throughout training in a new task. This introduces some correlation amongst the outputs, however, it dramatically cuts down on training time (another engineering consideration).

\vspace{-1.5ex}
\section{Evaluating the Attribute Classifier}
\vspace{-1.5ex}
As we get to a huge number of object categories (as in attribute classification), it is hard to estimate the number of categories that can be accurately categorized. Note that this is different than accuracy on a typical test set; this is because in real datasets, especially multi-label datasets, the distribution of categories of the instances heavily favours typical categories. Thus, standard accuracy scores do not capture how ``balanced'' the classifier is.

However, it is expensive to produce novel instances, especially from rare categories. Nevertheless, we don't need to test \emph{every} category to get an estimate of the number of categories that will be accurate. All we need is a distribution from which to draw instances; once we have this we can sample instances from this distribution and use this to produce a bound on the fraction of categories that are accurate. We can also use statistical tools to produce a confidence: the probability that our estimating technique will produce a correct bound (that is, a bound that includes that actual fraction of accurate categories).


Let $X$ be an instance space. We consider a set of possible categories
into which elements of an instance space can be classified which we
denote $C$. We consider a joint distribution of inputs $P(x|c$)
where $c\in C$ and $x$ is in the set of all instances in $X$ from
category $C$. When this instance is put into a classifier the output
is a category $\hat{c}\in C$ which is the classifier's estimate of
the instance's category $c$. We let $X_{c}$ (where $c\in C)$ represent
the set of instances that are in category $c$. 

Of all the $C$ possible classes (which is equal to $2^{K}$ in
the case of $K$ binary attributes), we let the vector $(p_{1},\ldots,p_{|C|})$
be the vector of true-positive accuracies for each of the $C$ categories.
Precisely the true-positive accuracy is defined as $P(\hat{c}=c|c)$,
the probability that the classifier's estimate is equal to class
$c$, conditioned on $c$ being the class.

\begin{defn} We say that a particular category $c$ is \emph{$\alpha$-accurate
}if the classifier has a true-positive accuracy for category $c$
greater than or equal to $\alpha$. Note that this is \emph{not }a
random variable, it is a \emph{fixed property of the category and
the classifier. }
\end{defn}

\begin{defn}We call the fraction of categories that are $\alpha$-accurate
the \emph{accurate fraction }and denote it $\theta$.
\end{defn}

Our goal is to estimate $\theta$. We do so by defining  a \emph{fraction-accurate estimator}, with parameters $\epsilon_{1}$,
$\epsilon_{2}$, $\alpha$, $M$ and $N$. In such an estimator, we first
randomly draw $N$ categories (each drawn from the set $C$ of categories, with replacement),
and for each of these $N$ categories produce $M$ sample instances
from those categories. We feed these instances into the classifier and compare the classifier's output to the true category, producing a vector of empirical accuracies,
which we denote $(\hat{p}_{1}\ldots\hat{p}_{N})$. We estimate that
all those values in this vector that are an $\epsilon_{1}$ amount
greater than or equal to $\alpha$ correspond to a \emph{classifiable}
category. We compute the fraction of these $N$ categories that are
classifiable, which we call $\hat{q}$, and then claim that: ``there
are at least $(\hat{q}-\epsilon_{2})$ categories that are classifiable.''
We call the value $\epsilon_{1}$ the \emph{accuracy deviation threshold}
and the value $\epsilon_{2}$ the \emph{fraction deviation threshold}. We call such an estimator a $(\alpha, N, M, \epsilon_1, \epsilon_2)$-fraction accurate estimator.

Note that the estimator is also associated with a \emph{random experiment,
}which in this case corresponds to randomly drawing $N$ categories
and then drawing $M$ instances of each category. Thus it assumes there is some distribution from which instances can be drawn.

We define
\begin{equation}
c_{p,n}(x)=\sum_{i=x}^{n}{n \choose i}p^{i}(1-p)^{n-i}\label{eq:cpnTaleOfTheBinomial}
\end{equation}
which is the the weight in the tail of a binomial distribution with
number of coin flips $n$ and coin bias $p$. In other words, it is
the probability that a coin with probability of heads $p$ comes up with $x$ or more
heads after $n$ flips.

\begin{theorem}\label{thm:confidence}The confidence of an $(\alpha, N, M, \epsilon_1, \epsilon_2)$ fraction-accurate estimator is bounded by:
\[
\mbox{confidence}\ge\min_{0<\theta<1}1-c_{c_{\alpha,M}\left(\left\lceil (\alpha+\epsilon_{1})M\right\rceil \right),N-\lfloor \theta+\epsilon_{2}/2)N \rfloor \rfloor}\left(\left\lceil \frac{\epsilon_{2}}{2}N\right\rceil \right)-c_{\theta,N}(\left\lceil (\theta+\epsilon_{2}/2)N\right\rceil )
\]
where we recall the definition in (\ref{eq:cpnTaleOfTheBinomial}), the weight of the tale of a binomial distribution.
\end{theorem}

\textbf{Proof:} Let the event $E(\theta,N,M,\epsilon_{1},\epsilon_{2})$
be the event that a random experiment followed by a fraction-accurate
estimator produces an estimate that does not include $\theta$ (that is, the event of an error). We
omit the arguments and denote this event $E$. Our goal shall be to
upper bound this probability and we do so using straightforward probability
arguments.

To bound this probability, we define the following events:

\[
A:\mbox{ \{the event that we draw greater than or equal to \ensuremath{(\theta+\epsilon_{2}/2)N} classifiable categories\}}
\]

\begin{align}
B:\mbox{ \{The event that we draw less than \ensuremath{(\theta+\ensuremath{\epsilon_{2}}/2)N} classifiable categories}\\
 \mbox{but more more \ensuremath{(\theta+\ensuremath{\epsilon_{2}})N}}\mbox{of them are classified as accurate}\}=\overline{A}\cap E
\end{align}
In the definition of $B$ we observe that the event that ``more than
$\ensuremath{(\theta+\ensuremath{\epsilon_{2}})N}$ of categories
are classified as accurate'' will result in a lower bound estimate
of $\theta$ at least slightly greater than $\theta$. In other words, this event is $\overline{A} \cap E$. Observe that
by using elementary set theory operations:

\[
E=(A\cap E)\cup(\overline{A}\cap E)\subseteq A\cup(\overline{A}\cap E)=A\cup B
\]

Thus we can conclude that:

\[
P(E)\le P(A\cup B)
\]

We now study these two events $A$ and $B$ separately.

We first bound event $A$, the
event that we draw greater than or equal to $(\theta+\epsilon_{2}/2)N$
classifiable categories when there are only $\theta$ classifiable
categories.

The probability can then be bounded by the tail of a binomial distribution:
\begin{equation}
P(A)\le\sum_{i=\left\lceil (\theta+\epsilon_{2}/2)N\right\rceil }^{N}{N \choose i}\theta^{i}(1-\theta)^{N-i}=c_{\theta,N}(\left\lceil (\theta+\epsilon_{2}/2)N\right\rceil )\label{eq:ProbOfA}
\end{equation}

Let's let $C$ be the event that a classifier with accuracy equal to $\alpha$
has empirical accuracy greater than $\alpha+\epsilon_{1}$. As a function of $M$,
the number of instances drawn of a particular category, we see that
this is related to the binomial distribution:
\begin{equation}
P(C)=\sum_{i=\left\lceil (\alpha+\epsilon_{1})M\right\rceil }^{M}{M \choose i}\alpha^{i}(1-\alpha)^{M-i}=c_{\alpha,M}\left(\left\lceil (\alpha+\epsilon_{1})M\right\rceil \right)\label{eq:probabilityOfNotAccurateClassifiedAsAccurate}
\end{equation}
where we use the definition of the function in (\ref{eq:cpnTaleOfTheBinomial}) to simplify the expression.
We now can use this expression to write an expression for the probability
of event $B$. Recall that event $B$ includes the event that $\frac{\ensuremath{\epsilon_{2}}}{2}N$
non-classifiable categories have empirical accuracy at least $\alpha+\epsilon_{1}$.

We note that the probability of the event that least $X$ of a set of categories with accuracy less than $\alpha$ have empirical accuracy over $\alpha+\epsilon$ is maximized when all the categories have  accuracy $\alpha$. We can prove this using a simple exchange argument.  We let $Q$ be the event that the number of categories with empirical accuracy above $\alpha+\epsilon_2$ exceeds $X$ for some arbitrary $X$. Suppose the accuracies were written $p_1, \ldots, p_j$. Then consider a particular category with accuracy $p_i<\alpha$. We let the event $i$ denote the event that this category is classified as accurate. We have:
\[P(Q)=P(Q|i)p_i+P(Q| \bar{i})(1-p_i)
\]
We see this expression increases with increasing $p_i$ because $P(Q| \bar{i})\le P(Q|i)$, because the event that we have at least $X$ categories being accurate conditioned on $i$ being inaccurate is strictly a subset of the event that we have at least $X$ categories accurate given $i$ being accurate. Thus this probability can be replaced with the value $\alpha$ and then the probability of $Q$ increases.
Thus:
\[
P(B)<P\mbox{\ensuremath{\left(\left\lceil \frac{\epsilon_{2}}{2}N\right\rceil \mbox{ of the at least \ensuremath{N-\lfloor \theta+\epsilon_{2}/2)N \rfloor} not accurate classifiers classified as accurate}\right)} }
\]
The probability that a particular not-accurate classifier gets classified
as accurate is bounded as in (\ref{eq:probabilityOfNotAccurateClassifiedAsAccurate}).
Thus:
\begin{eqnarray}
P(B)  <  \sum_{i=\left\lceil \frac{\epsilon_{2}}{2}N\right\rceil }^{N-\lfloor \theta+\epsilon_{2}/2)N \rfloor)N \rfloor}{N-\lfloor \theta+\epsilon_{2}/2)N \rfloor)N \rceil \choose i}\left(c_{\alpha,M}\left(\left\lceil (\alpha+\epsilon_{1})M\right\rceil \right)\right)^{i}\\
\left(1-c_{\alpha,M}\left(\left\lceil (\alpha+\epsilon)M\right\rceil \right)\right)^{(N-\lfloor \theta+\epsilon_{2}/2)N \rfloor)-i}\label{eq:probOfB}\\
 =  c_{c_{\alpha,M}\left(\left\lceil (\alpha+\epsilon_{1})M\right\rceil \right),N-\lfloor \theta+\epsilon_{2}/2)N \rfloor) \rceil}\left(\left\lceil \frac{\epsilon_{2}}{2}N\right\rceil \right)
\end{eqnarray}


We can now conclude by combining (\ref{eq:ProbOfA}) and (\ref{eq:probOfB})
we get:
\[
P(E)<c_{c_{\alpha,M}\left(\left\lceil (\alpha+\epsilon_{1})M\right\rceil \right), N-\lfloor \theta+\epsilon_{2}/2)N \rfloor }\left(\left\lceil \frac{\epsilon_{2}}{2}N\right\rceil \right)+c_{\theta,N}(\left\lceil (\theta+\epsilon_{2}/2)N\right\rceil )
\]
 and thus
\[
\mbox{confidence}\ge\min_{0<\theta<1}1-c_{c_{\alpha,M}\left(\left\lceil (\alpha+\epsilon_{1})M\right\rceil \right),N-\lfloor \theta+\epsilon_{2}/2)N \rfloor }\left(\left\lceil \frac{\epsilon_{2}}{2}N\right\rceil \right)-c_{\theta,N}(\left\lceil (\theta+\epsilon_{2}/2)N\right\rceil )
\]
We can estimate this value of
this over all valid $\theta$ by sweeping over the value of $\theta$
and choosing the minimum. Computationally this is not \emph{exactly} the minimum
but we increase the fineness of our sweep and show that the estimated
confidence changes only minimally.

\vspace{-1.5ex}
\section{Experiments}
\vspace{-1.5ex}

In this section we detail our experimental results. We test the naive repetition technique, and compare it to the parity partition coding technique. Consistent with the theoretical predictions, the parity partition technique consistently outperforms the naive repetition technique when the same number and size of models are used.

Models are trained to optimize the bit-accuracy of their respective codes (though we evaluate them using the $f1$ score of the decoded predictions), using binary cross-entropy as surrogate for continuous optimization. For the parity models, we train parity classifiers for all two-bit parity functions and use the code induced by appending these classifiers together. For all experiments, we used the Adam optimizer \cite{kingmaAdam} with a learning rate of 0.001 and a batch size of 64.

For our multi-MNIST classification model,  the architecture is composed of two feed-forward modules. First, we have six stacked ResNet blocks which act as feature extractors, then we first apply a $1\times 1$ convolution to reduce the number of channels followed by a global average pooling layer.  We then use BatchNorm before applying the quadratic transformation.  The final layer is a linear layer with sigmoid activation.
In Table \ref{tab:sharpness_mnist} we show an ablation study where we compared the baseline identity technique (which is just a $K$-output multi-target classifier) with the repetition technique (an ensemble of primitive models), as well as the parity technique. We test the weight-transfer techniques as well as the targeted bagging technique. 
\begin{table}[!h]
\centering
\begin{tabular}{|c|c |c|c|}
\hline
\multirow{2}{*}{Code} & Weight & \multicolumn{2}{c|}{Targeted Bagging}\\
\cline{3-4}
& transfer & no & yes\\
\hline
\multirow{2}{*}{Parity} & no & 	0.816 $\pm$ 0.011 / 0.377 $\pm$ 0.0471 & \textbf{0.847 $\pm$ 0.004} / \textbf{0.203 $\pm$ 0.0056} \\
\cline{2-4}
& yes & 0.766 $\pm$ 0.024 / 0.32 $\pm$  0.039 & 0.765 $\pm$ 0.025 / 0.321 $\pm$ 0.039\\
\hline
\multirow{2}{*}{Repetition} & no & 	0.703 $\pm$ 0.017 / 0.420 $\pm$  0.026 & 0.800 $\pm$ 0.014 /  0.261  0.018\\
\cline{2-4}
& yes & 0.724 $\pm$ 0.028 / 0.377 $\pm$ 0.047 & 0.731 $\pm$ 0.008 / 0.355 $\pm$ 0.038\\
\hline 
Identity & n/a & 0.736 $\pm$ 0.024 / 0.424 $\pm$  0.047& n/a\\
\hline
\end{tabular}
\caption{\label{tab:sharpness_mnist} Cells contain class $f_1$-scores (on the left) and average hamming distance (on the right) along with standard deviation for different codes on multi-MNIST, averaged over 10 trials.  The parity code incorporates 45 parity checks into the message string, corresponding to the XOR between all pairs of attributes. The repetition code uses 5 repetitions (to match the number of models in the parity code). The highest f1 score and lowest Hamming distance are emphasized in bold.}
\end{table}

\begin{figure}[!h]
    \centering
    \includegraphics[width=0.45\textwidth]{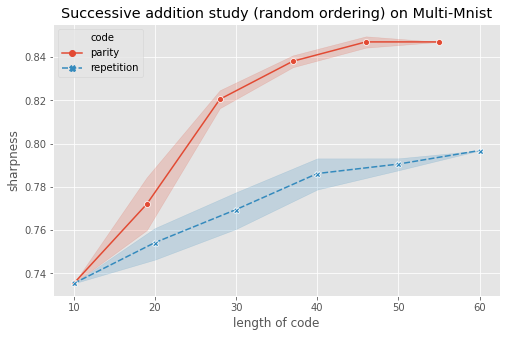}
    \caption{Effect of increasing number of classifiers for multi-MNIST classifier. For the repetition code, we train 6 separate models, each outputting the 10 primitive attributes. }\label{fig:sad}
\end{figure}

To demonstrate the effect of the gain from longer code sizes, we perform a successive addition study which we present in Figure~\ref{fig:sad}. In this experiment, we train multiple primitive attribute models, as well as multiple parity check models using quadratic feature transformation. We then evaluate the f1-score on these models when we aggregate them. We observe that as we the length of the associated codes increases, the parity technique outperforms the repetition technique.

\begin{figure}[!h]
    \centering
    \includegraphics[width=0.45\textwidth]{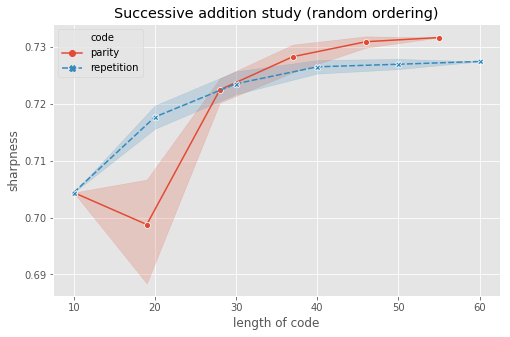}
    \caption{Effect of increasing number of classifiers for celebA classifier. Note that as we ensemble more models the accuracy of the parity technique outperforms the baseline repetition technique.}\label{fig:celebAsad}
\end{figure}

For our next set of experiments, we trained a multi-attribute facial feature recognizer trained on celebA with the following attribute template: \{Wearing necktie, Male, Gray hair, Chubby, Wearing hat, Blond hair, Bald, Heavy makeup, No beard, Eyeglasses\}. As this is a more computationally intensive task, we used the study presented in Table~\ref{tab:sharpness_mnist} to inform our choice of techniques used for celebA. Thus, for celebA, we use quadratic feature transformation (which trained faster in multiMNIST), split the dataset using targetted bagging, and do not use weight transfer. For our CelebA classification model, we use the same strategy as we did for the MultiMnist, but use a larger ResNet encoder as found in \cite{senerStateOfArt}, since it achieves state-of-the-art results.  The primitive attributes in the CelebA dataset are noticeably less balanced than in the MultiMnist case, (and thus, so are derived attributes), which we combat by adding frequency-weights to the binary cross entropy.  Unlike in the MultiMnist case, where targets were chunked to save on computational resources, for CelebA we train separate models for each attribute (primitive or derived).

Correct classification occurs when \emph{all} of these attributes are classified correctly (which explains the relatively low classification accuracy). Nonetheless, the parity partition technique outperforms the baseline identity techniques, and repetition techniques. We present these results in Figure~\ref{fig:celebAsad} and Table~\ref{tab:metrics:celeba}. We also compare the bit-level accuracies of the models for both multiMNIST and celebA. For celebA we see that our models have bit accuracies that exceed state of the art \cite{senerStateOfArt} for $8$ out of the $10$ attributes, and these results are shown in the appendix.

To utilize our fraction-accurate estimation technique, we employ two studies, one each for the multi-MNIST dataset and the celebA dataset. For the multi-MNIST dataset, we sample from all $2^{10}$ categories to estimate the fraction of categories that are accurate. We set as a threshold $\alpha=0.5$ to estimate the fraction of categories with accuracy over $0.5$. We choose accuracy deviation threshold $\epsilon_1=0.19$ and the fraction deviation threshold $\epsilon_2=0.19$, number of categories sampled $N=100$, and number of samples from each category $M=20$. We use Theorem~\ref{thm:confidence} to compute our confidence is $0.963$, and our study gives us a lower bound on the fraction of accurate categories as $0.46$. Thus, for our baseline model, with $96.3\%$ confidence we are sure that $46\%$ of the categories are accurate. On the other hand, when we use parity partition coding and the same set of parameters, we find that $74\%$ of the categories are accurate. For our repetition baseline, we get $64\%$. 

For celebA, we face a different issue compared to multiMNIST, because we cannot necessarily produce instances from all $2^{10}$ categories. However, we select a fraction of categories for which our dataset has multiple instances and sample from this set. We set accuracy threshold to $\alpha=0.1$, $N=100$, $M=10$, and $\epsilon_1=\epsilon_2=0.2$, for which we compute a confidence of $97.1\%$. For this set of parameters, for our baseline model we estimate there are at least $12\%$ accurate categories, for our repetition model, we estimate $35\%$, and for our parity model we estimate $27\%$ (less than the repetition model, but with empirical estimate within the fraction deviation threshold of the repetition model).

\begin{table}[!h]
    \centering
    \begin{tabular}{|c|c|c|}
        \hline
         Code & $f_1$ score & Hamming Distance   \\
         \hline
         Identity & 0.704 &0.346 $\pm$ 0.581 \\
         \hline
         Repetition & 0.726 & 0.316 $\pm$ 0.555 \\
         \hline
         Parity & \textbf{0.732} & \textbf{0.308 $\pm$ 0.550} \\
        \hline 
    \end{tabular}
    \caption{The f1 score and average hamming distance from ground truth label, for the celebA dataset classification challenge. We note that the parity model has a higher f1 score and lower hamming distance than the baseline repetition and identity models.}
    \label{tab:metrics:celeba}
\end{table}

\vspace{-1.5ex}
\section{Conclusion}
\vspace{-1.5ex}
We introduce parity partition coding, and argue using results from coding theory that this technique will result in an $O(K)$ savings in number of binary classifiers required to get high accuracy, where $K$ is the number of attributes. We test the technique on multiMNIST, a synthetic multi-label dataset, and a label classification challenge based on celebA, where we reach comparable to state of the art. We introduce the notion of sharpness, and show how to bound this quantity by sampling over categories while also producing a confidence estimate for this bound.

\bibliographystyle{unsrt}
\bibliography{bibtextDoc}

\newpage
\section{Appendix: Bit Accuracies}

\begin{table}[!h]
    \centering
    \begin{tabular}{|c|c|c|c|c|}
        \hline
         Attribute
         & \multicolumn{1}{|p{2cm}|}{\centering Sender \\ et. al}
         & \multicolumn{1}{|p{2cm}|}{\centering Baseline \\ accuracy} 
        & \multicolumn{1}{|p{2cm}|}{\centering Repetition-corrected \\ accuracy}
        & \multicolumn{1}{|p{2cm}|}{\centering Parity-corrected \\ accuracy} \\
         \hline
         Wearing Necktie & 0.965 & 0.965 & 0.970 & \textbf{0.971} \\
         \hline
        Male & \textbf{0.986} & 0.977& 0.982 & 0.982 \\
         \hline
        Gray Hair & 0.978 & 0.981 & 0.981 & \textbf{0.983} \\
         \hline
        Chubby & 0.955 & 0.952 & 0.957 & \textbf{0.957} \\
         \hline
        Wearing Hat & 0.989 & 0.990 & \textbf{0.991} & 0.990 \\
         \hline
        Blond Hair & 0.954 & 0.954& \textbf{0.959} & 0.958 \\
         \hline
        Bald & 0.989 & 0.986 & \textbf{0.989} & 0.989 \\
         \hline
        Heavy Makeup & 0.922 & 0.904& \textbf{0.912} & 0.910 \\
         \hline
        No Beard & \textbf{0.958} &  0.953 & 0.947 & 0.957 \\
         \hline
        Eyeglasses & 0.994 & 0.991 & \textbf{0.996} & 0.995 \\
         \hline
    \end{tabular}
    \caption{The accuracies of each individual attribute for our celebA classifiers. On this metric we see that our parity models and repetition models are comparable, and on these individual labels compare to state of the art. Bold indicates best value}
    \label{tab:accs:celeba}
\end{table}

\begin{table}[!h]
    \centering
    \begin{tabular}{|c|c|c|c|}
        \hline
         Attribute
         & \multicolumn{1}{|p{2cm}|}{\centering Baseline \\ accuracy}
        & \multicolumn{1}{|p{2cm}|}{\centering Repetition-corrected \\ accuracy}
        & \multicolumn{1}{|p{2cm}|}{\centering Parity-corrected \\ accuracy} \\
         \hline
0 & 0.981 $\pm$ 0.004 & 0.988 $\pm$ 0.001 & \textbf{0.991 $\pm$ 0.001} \\ 
\hline
1 & 0.943 $\pm$ 0.015 & 0.963 $\pm$ 0.007 & \textbf{0.974 $\pm$ 0.003} \\ 
\hline
2 & 0.966 $\pm$ 0.012 & 0.982 $\pm$ 0.004 & \textbf{0.986 $\pm$ 0.002} \\ 
\hline
3 & 0.969 $\pm$ 0.008 & 0.981 $\pm$ 0.003 & \textbf{0.984 $\pm$ 0.002} \\ 
\hline
4 & 0.963 $\pm$ 0.008 & 0.978 $\pm$ 0.004 & \textbf{0.982 $\pm$ 0.001} \\ 
\hline
5 & 0.958 $\pm$ 0.010 & 0.977 $\pm$ 0.001 & \textbf{0.981 $\pm$ 0.001} \\ 
\hline
6 & 0.954 $\pm$ 0.024 & 0.977 $\pm$ 0.003 &\textbf{ 0.983 $\pm$ 0.002} \\ 
\hline
7 & 0.944 $\pm$ 0.013 & 0.961 $\pm$ 0.005 & \textbf{0.969 $\pm$ 0.001} \\ 
\hline
8 & 0.962 $\pm$ 0.010 & 0.977 $\pm$ 0.002 & \textbf{0.980 $\pm$ 0.002} \\ 
\hline
9 & 0.936 $\pm$ 0.008 & 0.955 $\pm$ 0.007 & \textbf{0.968 $\pm$ 0.002} \\ 
\hline
    \end{tabular}
    \caption{The accuracies of each individual attribute for our MultiMnist classifiers. Bold indicates best value.  We see that parity decoding leads to the most accurate bits.}
    \label{tab:accs:celeba}
\end{table}
  
\end{document}